\documentclass[11pt]{article}

\usepackage[final]{acl}

\usepackage{booktabs}
\usepackage{multirow}
\usepackage{tabularx}

\usepackage{times}
\usepackage{latexsym}
\usepackage{amsmath}
\usepackage{amssymb}
\usepackage{url}

\usepackage[T1]{fontenc}

\usepackage[utf8]{inputenc}

\usepackage{microtype}

\usepackage{inconsolata}

\usepackage{graphicx}

\title{The Fragility and Dynamics of LLM Pruning-Recovery Steps}

\author{
 \textbf{Luis Frentzen Salim\textsuperscript{1,2}},
 \textbf{Komang Rinartha\textsuperscript{1}}
 \textbf{Chia-Chun Wu\textsuperscript{1}},
 \textbf{Tran Van Nhiem\textsuperscript{3}},
 \textbf{Lun-Wei Ku\textsuperscript{1}},
 \textbf{Yung-Hui Li\textsuperscript{3}},
\\
 \textsuperscript{1}Institute of Information Science, Academia Sinica,
\\
 \textsuperscript{2}National Taiwan University of Science and Technology,
\\
 \textsuperscript{3}AI Research Center, Hon Hai Research Institute,
\\
 \small{
   \textbf{Correspondence:} \href{mailto:luisfrentzen@gmail.com}{luisfrentzen@gmail.com}
 }
}

\begin{document}
\maketitle
\begin{abstract}
Structural pruning followed by a recovery stage is the standard recipe for compressing large language models (LLMs), yet the two stages are typically designed and tuned in isolation. In this work, we study how their interaction shapes the compressed model, sweeping ordering, the data that drives each stage, and the alignment of their training objectives. We find that instruction-following is the first capability to collapse under pruning, well before knowledge, reasoning, or core language modeling, and that much of the pruning in-domain loss is likely not knowledge loss but a domain-agnostic failure to respond coherently to the prompt. We find in our setting that in-domain data is the strongest pruning signal, whereas instruction-style data recovers the most. Stage order matters little, but only marginally. We further analyze how the importance ranking drifts before and after pruning. Together these results give practitioners a concrete recipe for composing pruning and recovery.
\end{abstract}

\section{Introduction}
\label{sec:intro}
Large Language Models (LLM) are useful in many contexts and applications ~\cite{NEURIPS2020_1457c0d6,
grattafiori2024llama3herdmodels,
shi2022languagemodelsmultilingualchainofthought}, however, their memory footprint is problematic for many deployment setting. This motivates structural pruning algorithm to reduce the memory footprint of LLMs. Structural pruning works by removing a cohesive structure altogether to reduce the parameters based on a pruning objective \cite{llmpruner, frantar2023sparsegptmassivelanguagemodels}. It is often followed by an intermediate recovery step by knowledge distillation (KD) \cite{hinton2015distilling} or supervised finetuning (SFT) \cite{wei2022finetunedlanguagemodelszeroshot} to recover the performance lost during removal of neural parameters. This framework to prune-then-recover has been widely adopted for deployment and practical usecases.

Despite their coupled nature, recovery and pruning are often treated as two distinct components. Yet, the success of one component may depend on the effectiveness of the other. In this study we study the relationship and alignment between structural pruning and performance recovery. Offering a practical recipe to prune and recover LLMs. Our study demonstrates that:
\begin{itemize}
    \item \textbf{Instruction following capabilities is the most fragile} as it degrades the fastest as you prune more parameters.
    \item \textbf{In-domain pruning and instruction following recovery works best}, we show that the degradation may not be attributed purely to knowledge loss but also due to the model losing the ability to respond coherently.
    \item \textbf{Loss matching yields no benefit}, aligning the recovery loss with the pruning loss does not offer significant benefit.
    \item \textbf{Important weights are likely to stay after pruning}, we show that top and bottom 20\% weights are likely to stay on their ranks, but rank retention went down as we increase pruning ratio, which could explain how iterative pruning works to preserve performance.
\end{itemize}

\section{Related Works}
\label{sec:related}
\begin{figure*}[t]
  \includegraphics[width=\textwidth]{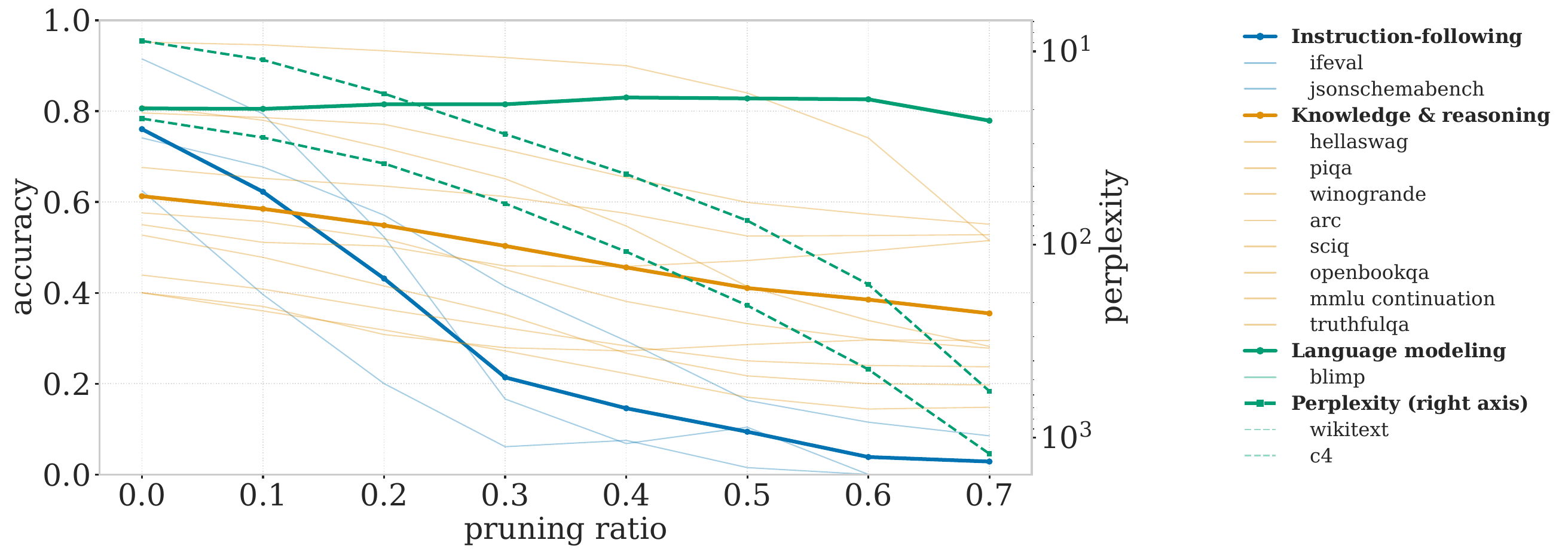}
  \caption{Performance over pruning ratios for Llama 3.1 8B Instruct. Instruction following degrades earlier.}
  \label{fig:fragilty}
\end{figure*}

\subsection{Language Model Pruning}
Pruning reduces model size by removing parameters and is commonly
divided into unstructured pruning, which zeroes out individual weights, and structured pruning, which removes entire coherent units and is therefore more hardware-friendly without
specialized sparse kernels~\cite{li2017pruning}.
~\citet{llmpruner} shows task-agnostic structured pruning by building a dependency graph that groups coupled structures, estimating group importance from first-order gradient information. Sheared-LLaMA~\cite{xia2024sheared} prunes a larger model to a target shape via targeted structured pruning and dynamic batch loading.

\subsection{Calibration and Importance Estimation}
Pruning decisions are commonly guided by estimates of how much the model loss would change if a parameter, neuron, head, or structured block were removed. \cite{NIPS1989_6c9882bb} shows that second-order derivative information can be used to compute parameter saliency for network pruning, and \cite{NIPS1992_303ed4c6} shows how fuller second-order information can improve pruning decisions by accounting for interactions between parameters. \cite{molchanov2017pruningconvolutionalneuralnetworks} shows that a first-order Taylor expansion of the loss can serve as an effective criterion for removing filters while preserving task performance. In LLMs pruning methods often estimate importance from a calibration set. Work like \cite{williams-aletras-2024-impact, bandari-etal-2024-c4} shows how downstream performance after post-training quantization and pruning can vary substantially depending on the calibration data, and \cite{ji2025bewarecalibrationdatapruning} shows that calibration data similar to the model’s training distribution can improve pruning performance.

\subsection{Performance Recovery}
Since pruning inherently causes performance loss, most pruning pipelines include a recovery stage to regain lost performance. Recovery can be viewed through several objective families. Generally, KD or SFT are common types of recovery. Several LLM pruning methods rely on post-pruning recovery. \cite{llmpruner} shows that structurally pruned LLMs can recover performance through lightweight LoRA tuning, and \cite{zhang-etal-2024-loraprune} shows that LoRA-guided pruning can be a memory-efficient way to prune models. While \cite{xia2024sheared} shows that structured pruning followed by continued pretraining can produce competitive smaller LLMs with substantially less compute than training from scratch. More recent work further emphasizes recovery design. Such as self-distillation recovery \cite{thangarasa2025selfdatadistillationrecoveringquality}, and \cite{liu2025patpruningawaretuninglarge} introducing sparsification during recovery.

\section{Methodology}
\label{sec:methodology}
\begin{figure*}[t]
    \includegraphics[width=\textwidth]{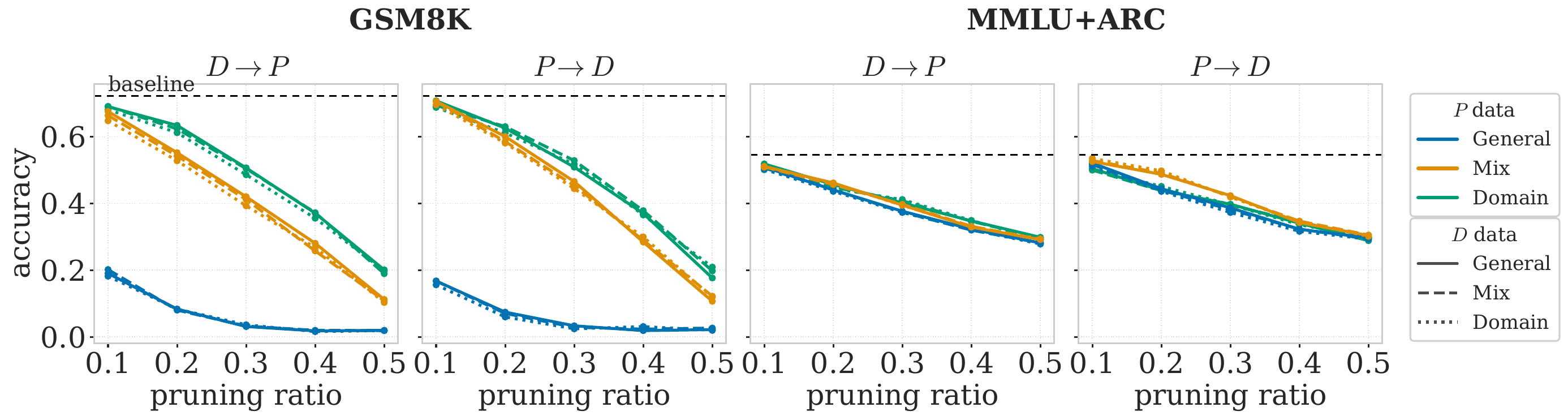}
    \caption{Performance on the domain and general tasks over pruning $r$, with one line for each configuration.}
\end{figure*}

\subsection{Structured Pruning}

We adopt the method of~\cite{llmpruner} as our pruning backbone, using a first-order Taylor criterion to approximate the importance of each structure. Given a calibration set $\mathcal{C}$, we estimate the importance of a weight element $W_i^k$ by the change in loss incurred when it is zeroed. A first-order Taylor expansion approximates this change by the element's gradient--weight product, giving

\begin{equation}
    I_{W_i^k} \approx
    \left| \frac{\partial \mathcal{L}(\mathcal{C})}{\partial W_i^k}\, W_i^k \right|,
    \label{eq:first-order}
\end{equation}

which we evaluate with a single backward pass of the causal language modeling (CLM) loss. Higher-order terms are dropped, as forming the full Hessian is infeasible at LLM scale. We follow the \emph{block-wise} setting, in which functionally coupled weights are grouped and removed together, so the unit of pruning is an attention head or an MLP intermediate channel. The element scores within a group $\mathcal{G}$ are aggregated by summation,

\begin{equation}
    I_{\mathcal{G}} = \sum_{k \in \mathcal{G}} I_{W_i^k},
    \label{eq:group}
\end{equation}

and groups are ranked locally within each block.

\subsection{Models}

The primary model used in this study is Llama 3.2 3B Instruct and Llama 3.1 8B Instruct \cite{grattafiori2024llama3herdmodels}. We fully utilize the instruction chat template provided with the model in all evaluation and training unless specified otherwise. For the KD recovery stage we use Gemma 3 12B Instruct~\citep{gemma3} as the teacher model.

\subsection{Pipeline Ablation}

To find the optimal configuration of the pruning-recovery pipeline, we run an ablation following the permutation of every component and its respective calibration set. We apply structural pruning and a KD recovery stage, which from this section onward will also be referred to as $P$ and $D$, respectively. We sweep over three factors that define how these operations are composed. The first is the \emph{order} of the two stages, either $D$-first or $P$-first. The second is the pruning ratio $r \in \{0.1, 0.2, 0.3, 0.4, 0.5\}$. The third is the \emph{kind} of data in each stage. We employ a \textsc{General} (pretraining-like) set, the in-\textsc{Domain} set, or a \textsc{Mix} of the two.

We use 32 examples from each dataset to prune, and we sample 1.5 million tokens from each dataset for training to normalize calibration size. The datasets used as calibration in this experiment are:

\begin{itemize}
    \item BookCorpus~\citep{Zhu_2015_ICCV} is a large collection of unlabelled text drawn from thousands of freely available books. We use it as the \textsc{General} calibration corpus for pruning. 
    \item Alpaca~\citep{alpaca} is a corpus of $52$K instruction-following demonstrations covering a broad range of everyday tasks. Its general, instruction-shaped content makes it our \textsc{General} source for the recovery stage. This data is used for our KD setup to match GOLD trainer's chat format requirement and to induce instruction-following capabilities.
    \item GSM8K~\citep{cobbe2021gsm8k} is a dataset of roughly $8.5$K grade-school math word problems. It is the in-\textsc{Domain} dataset in our study. We sampled only from the training set for recovery and used the test set for evaluation.
\end{itemize}

\subsection{Performance Recovery}

We recover pruned models with KD, specifically the General Online Logit Distillation (GOLD) trainer with the trl library~\cite{vonwerra2020trl}. GOLD performs \emph{on-policy} KD, the student generates its own continuations and is trained to match the teacher's token distributions on those self-generated sequences. We use the Gemma 12B model to provide a teacher with significantly larger size compared to the Llama models and the cross-tokenizer setting is enabled by GOLD. All recovery is full-parameter finetuning. Complete hyperparameters are listed in Appendix~\ref{app:hyperparams}.

\begin{figure*}[t]
  \includegraphics[width=\textwidth]{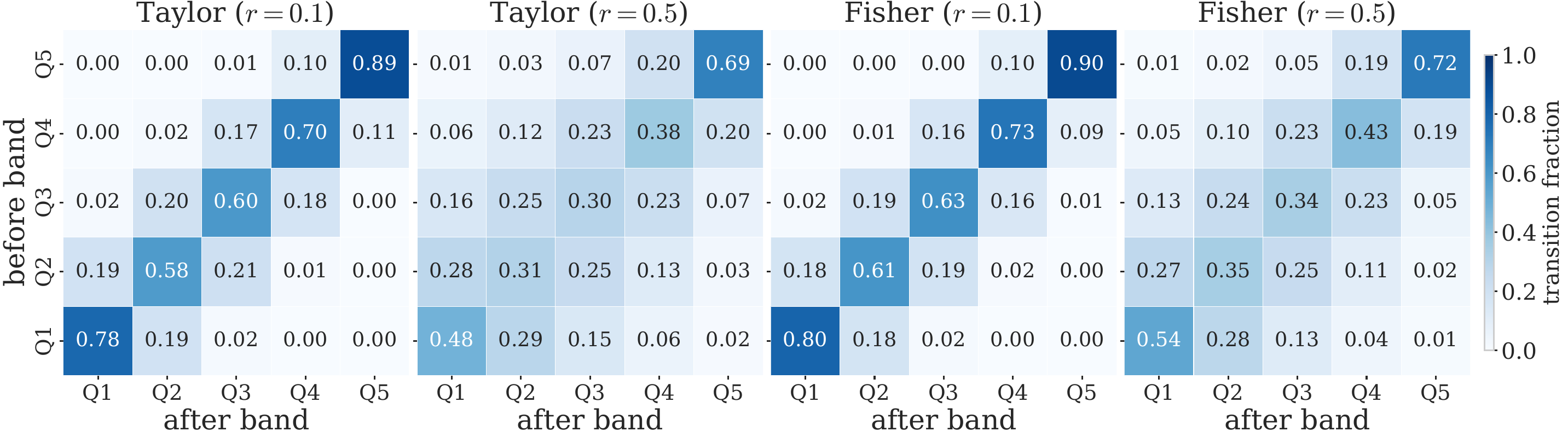}
  \caption{Spearman correlation of moving weight rankings before and after pruning with $r$ $0.1$ and $0.5$. It is observed that the most and least important weights tend to stay as the model is pruned.}
  \label{fig:spearman_movement}
\end{figure*}

\subsection{Loss Alignment}

To isolate whether matching the recovery loss to the pruning loss is beneficial (\S\ref{sec:alignment}), we run a controlled comparison with the \textsc{General} dataset. We vary only the loss used in each stage. Recovery is performed either by SFT under the CLM loss or by KD under the GOLD loss, and pruning importance is correspondingly estimated under the matching loss. A configuration is \emph{aligned} when both stages optimize the same loss and \emph{misaligned} otherwise. In this analysis specifically, both the pruning and the recovery losses are computed \emph{off-policy} and training budget is set to 2.5 million tokens. Hyperparameters are listed in Appendix~\ref{app:hyperparams}.

\subsection{Fisher-based Importance Scoring}

For the structural importance analysis, we additionally score each structure with the diagonal empirical Fisher information, a curvature-based criterion that depends only on the gradient. Replacing the intractable Hessian diagonal with the expected squared gradient~\cite{hassibi1992obs}, the importance of the $k$-th weight element $W_i^k$ is

\begin{equation}
    I^{\mathrm{F}}_{W_i^k} \approx \frac{1}{N} \sum_{j=1}^{N}
    \left( \frac{\partial \mathcal{L}(\mathcal{C}_j)}{\partial W_i^k} \right)^2,
    \label{eq:fisher}
\end{equation}

i.e. the mean squared gradient over the $N$ calibration samples. As with the Taylor criterion, element scores are summed within each structure to obtain its group importance $I^{\mathrm{F}}_{\mathcal{G}} = \sum_{k \in \mathcal{G}} I^{\mathrm{F}}_{W_i^k}$. Unlike Equation~\ref{eq:first-order}, this score weights elements purely by gradient curvature and are independent of the weight magnitude.

\subsection{Structural Importance Analysis}

To assess how pruning reshapes the importance and loss landscape, we compare per-structure importance scores of the dense model against its pruned counterpart. For a given pruning configuration, we score every structure of both the unpruned and the pruned model on the same calibration set $\mathcal{C}$, using both the first-order Taylor criterion (Equation~\ref{eq:first-order}) and the empirical Fisher criterion (Equation~\ref{eq:fisher}). Each surviving structure of the pruned model is mapped back to its original position in the dense model through a structure map recorded at pruning time. Because importance magnitudes differ systematically across depth, all comparisons are performed within a layer and then aggregated across layers, weighted by the number of surviving structures. For the rank-mobility analysis, we partition structures within each layer into five percentile bands (quintiles, $Q_1$--$Q_5$) by importance, and track how survivors move between bands from the dense to the pruned model. We repeat the analysis for every pruning ratio $r$ and for both criteria.

\subsection{Experiment Details}

All experiments are done with 8 NVIDIA L40 GPUs. We run each training once with a fixed seed with the trl library for GOLD and unsloth \cite{unsloth} for SFT.

\subsection{Evaluation}
\label{sec:eval}

We run all evaluations with a fixed random seed using the LM Evaluation Harness library \cite{eval-harness}. We evaluate our model checkpoints on three main tasks:

\begin{itemize}
    \item \textbf{GSM8K} \cite{cobbe2021gsm8k} grade-school math word problems requiring arithmetic reasoning. It serves as our in-domain task and probes mathematical reasoning.
    \item \textbf{MMLU} \cite{hendrycks2021measuringmassivemultitasklanguage} multiple-choice questions spanning 57 subjects, probing broad, general-purpose world knowledge.
    \item \textbf{ARC} \cite{clark2018arc} grade-school science questions (Easy and Challenge splits) that probe scientific knowledge and reasoning. MMLU and ARC will also be referred to as General in later sections for brevity.
\end{itemize}

To further assess the performance of pruned models on a diverse range of tasks, we use a broader benchmark suite, grouped into three capability groups. We report accuracy for all tasks except WikiText and C4, for which we report perplexity.

\begin{itemize}
    \item \textbf{Instruction-Following} probes whether the model can still parse and obey explicit instructions and emit well-formed structured output. Consists of IFEval \cite{zhou2023instructionfollowingevaluationlargelanguage} and JSONSchemaBench \cite{geng2025jsonschemabench}, measuring instruction-level adherence and the validity of structured output.
    \item \textbf{Knowledge \& Reasoning} probes the world knowledge and commonsense, scientific, and factual reasoning measured through question answering. Comprises of HellaSwag \cite{zellers2019hellaswag}, PIQA \cite{Bisk2020}, WinoGrande \cite{sakaguchi2019winograndeadversarialwinogradschema}, ARC-Easy and ARC-Challenge \cite{clark2018arc}, SciQ \cite{welbl-etal-2017-crowdsourcing}, OpenBookQA \cite{mihaylov-etal-2018-suit}, MMLU \cite{hendrycks2021measuringmassivemultitasklanguage}, and TruthfulQA \cite{lin-etal-2022-truthfulqa}.
    \item \textbf{Language Modeling} probes the model's core modeling performance. Consists of BLiMP \cite{warstadt2020blimp}, WikiText \cite{merity2016pointer} and C4 \cite{JMLR:v21:20-074}.
\end{itemize}

\section{Results and Discussion}
\label{sec:discussion}
\begin{figure}[t]
  \includegraphics[width=\columnwidth]{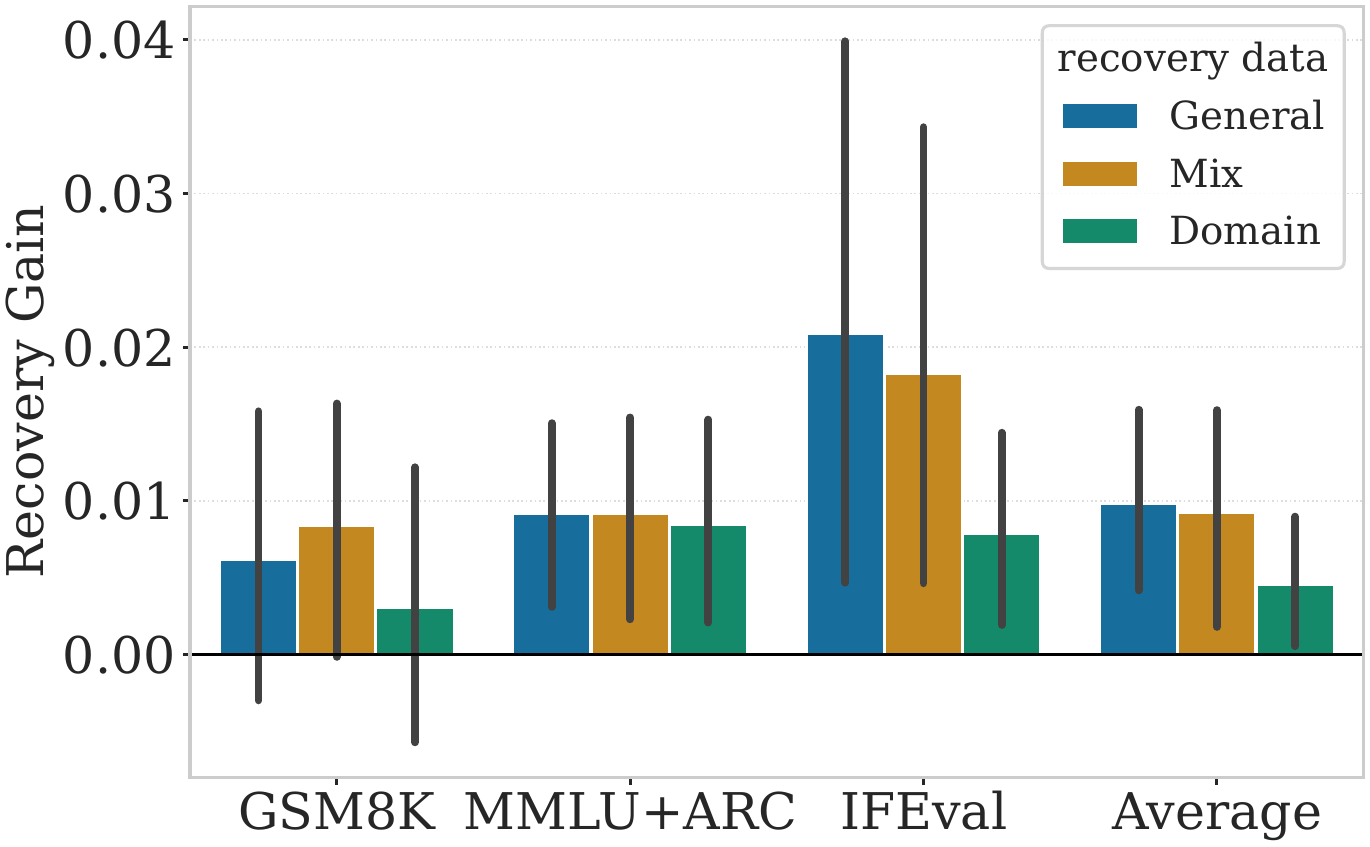}
  \caption{Recovery gain by data, for $P$-first ($P\rightarrow D$) sequences. Each bar is the change in a metric relative to the matched pruning-only model with no recovery, averaged over pruning corpora and ratios.}
  \label{fig:recovery}
\end{figure}

\subsection{Instruction-following goes first}
\label{sec:fragility}

To identify which capabilities are most vulnerable to compression, we evaluate Llama 3.1 8B Instruct model over ratio from $0.1$ to $0.7$. The model is then evaluated over three task categories shown in \S\ref{sec:eval}. We observe that tasks performance degrade at sharply different rates and in a consistent order. \emph{Instruction-following collapses first}, by $30\%$ sparsity IFEval has fallen from $0.74$ to $0.41$ and structured-output adherence (JSONSchema) has largely collapsed. Knowledge and reasoning degrade more gracefully, declining roughly linearly and retaining over half the original accuracy at $50\%$ sparsity while language modeling task declines last. 

We observe that perplexity-based metrics are not a reliable measure of quality for compressed models, since they capture token-level likelihood and not whether the model outputs coherently. This should inform how we approach recovery, which is commonly carried out with knowledge- or domain-specific data rather than instruction-tuning data. Much of the degradation observed across downstream tasks may be attributable not to knowledge or specific domain ability loss but to the model's failure to answer in the expected form or to answer at all due to its compromised instruction-following. An extensive output analysis and supportive proofs present in Appendix~\ref{app:response-failure}. Accordingly, replaying instruction-tuning data during recovery could be a natural strategy to mitigate this loss.

\begin{table}[t]
\centering
\begin{tabular}{lcc}
\toprule
Pruning corpus & $\Delta$ & $p$ \\
\midrule
All           & $-0.009$ & ${<}0.001$ \\
Single-source & $-0.003$ & $0.22$ \\
Mixed         & $-0.020$ & ${<}0.001$ \\
\bottomrule
\end{tabular}
\caption{Pipeline ordering result averaged over tasks, order and pruning ratios. $\Delta=(\text{$D$-first})-(\text{$P$-first})$, negative values favor pruning-first.}
\label{tab:order}
\end{table}

\subsection{Instruction recovery helps most}
\label{sec:recovery}

\S\ref{sec:fragility} showed that instruction-following is the first to fall under pruning. We now ask whether it is also the deficit that recovery should prioritize. For every pruning-first configuration in our ordering ablation we measure the \emph{recovery gain}, the change in each metric relative to the matched pruning-only model, and group it by the data used for recovery, holding the pruning corpus and ratio fixed (Figure~\ref{fig:recovery}).

Across all three recovery sets, general capability and instruction-following improve, and the bulk of the recoverable gain is concentrated in instruction-following. General- and Mix-recovery improve the overall average significantly ($+0.010$, $p=0.011$ and $+0.009$, $p=0.022$), while domain-only yields only a marginal gain ($+0.005$, $p=0.06$). General and Mixed recovery shows a nominally significant advantage over Domain recovery, even for in-domain evaluation. This suggests an asymmetry between structure selection and weight refitting. In-domain or mixed data provides the strongest pruning signal, whereas instruction-style data provides the strongest recovery signal.

\subsection{Pruning-first helps only modestly}
\label{sec:order}

In practice, we prune first to then recover loss performance. We test whether the order matters at all by comparing distill-first and pruning-first across all configurations as shown in Table~\ref{tab:order}. Pruning-first is consistently better on the overall average, winning $31$ of $45$ matched pairs. The effect is statistically reliable but small. It is also concentrated in the mixed-corpus case ($p<0.001$), for a single-source pruning corpus the two orders are statistically indistinguishable ($p=0.22$). While the result confirms the norm, we find minimal difference between the distill-first and pruning-first pairs. Although pruning-first is better on average, the effect is small outside the mixed-corpus condition. This suggests that distill-first may still be an option when post-pruning recovery is difficult (e.g., small model with huge pruning ratio), though we leave extensive testing for future work.

\subsection{Loss matching gives no detectable benefit}
\label{sec:alignment}

\begin{table}[t]
\centering
\begin{tabular}{lccc}
\toprule
Metric & Aligned & Misaligned & $p$ \\
\midrule
GSM8K    & \textbf{0.075} & \textbf{0.075} & 0.98 \\
General & 0.379 & \textbf{0.385} & 0.13 \\
Average  & 0.387 & \textbf{0.393} & 0.12 \\
\bottomrule
\end{tabular}
\caption{Alignment performances over pruning $r$.}
\label{tab:alignment}
\end{table}

The first-order Taylor criterion approximates the loss change $\Delta\mathcal{L}$ from removing a structure, so one might expect that pruning under a given objective would be better suited for recovery under the same objective. Empirically, matching the recovery loss to the pruning loss yields no benefit. Across all three metrics, the matched configurations are statistically indistinguishable from the mismatched ones (Table~\ref{tab:alignment}). If anything, it slightly favors the \emph{mismatched} setting on general tasks and on the overall average. This is consistent with the broader observation from prior work that static importance estimates can become stale once parameters are subsequently optimized~\citep{sanh2020movement}. Within our tested losses and recovery budget, the choice of matched versus mismatched recovery loss appears to have little measurable effect. Furthermore, both losses used in this study may carry distinct optimization characteristics. While we measure the effect of matching the two stages, the alignment is not fully disentangled from these differences. We leave further exploration to future work.

\subsection{Stable extremes, mobile middle}

To understand how pruning affects the model's weights, we track how the importance ranking of the surviving structures changes between the dense model and the pruned model. For each configuration, we recompute per-structure importance on the pruned network, map every survivor back to its position in the dense model, and compare the two rankings within each layer (Figure~\ref{fig:spearman_movement}).

The ranking movement is minimal under small pruning $r$ but increases as the $r$ grows. The rank correlation between the baseline and pruned importance falls from $0.95$--$0.98$ at $10\%$ sparsity to $0.69$--$0.77$ at $50\%$. Among surviving structures, the extremes of the importance ranks are most stable, while the mid-ranked moves the most, retention distribution is U-shaped. The pattern holds under the Taylor and Fisher criteria. Given this context, matching the losses may align to a stale signal, suggesting that the calibration set and objective are clearer signals towards the eventual model profile. These findings could also explain why iterative pruning works well to retain model's performance rather than one-shot pruning. 

\section{Conclusion}
\label{sec:conclusion}

We study the properties of the pruning-recovery steps and how the alignment between the two components affects overall performance. Pruning imprints the model's capability profile at selection time and, in our setting, favors in-domain or mixed data. We observe that instruction-following degrades the fastest under pruning, and that the resulting performance loss is not purely a loss of domain knowledge but is driven in large part by the degradation of the model ability to follow instructions. Recovery, in turn, is most effective when it replays instruction-style data, restoring the instruction-following ability that degrades the most. Stage order matters little in absolute terms, pruning-first wins consistently but by under a point. Finally, the importance ranking shifts more as the pruning ratio increases. Taken together, these results offer practical insights for composing the pruning-recovery stages more effectively and to better understsand the mechanism behind the pipeline.

\section{Future Works}
\label{sec:conclusion}
Further exploration on how the loss function affects the two stages is a promising future direction to better understand the alignment between pruning and recovery objective. Furthermore, understanding why some capabilities fail earlier than the other and whether the ranking shift during pruning may have any correlations to performance is an open question to study in the future.

\section{Limitations}
\label{sec:limitation}
Our pipeline ablation use a single model family and scale and a single in-domain task due to compute limitations. Our loss-alignment comparison does not fully separate the effect of matching the two objectives from their differing optimization characteristics. Lastly, our trainings are done within limited tokens budgets and we did not train further due to compute limitation.

\bibliography{custom,anthology-1,anthology-2}

\clearpage
\appendix
\section*{Appendix}
\label{sec:appendix}





\subsection{Hyperparameters}
\label{app:hyperparams}

We list the hyperparameters for the pruning and recovery stages below. Any
setting not listed is left at its library default (trl for distillation and supervised fine-tuning). All evaluations use the
default settings of the LM Evaluation Harness with chat template enabled unless stated otherwise. All other non-ablated values are held fixed across runs.

\begin{table}[h]
\centering
\begin{tabular}{lll}
\toprule
Hyperparameter & GOLD & SFT \\
\midrule
Precision              & bf16           & bf16 \\
Max length    & $2048$         & $2048$ \\
batch size  & $2$            & $4$ \\
Gradient accumulation  & $16$           & $8$ \\
Global batch size      & $256$          & $32$ \\
Learning rate          & $1\times10^{-6}$ & $1\times10^{-6}$ \\
Epochs                 & $1$            & $1$ \\
Warmup ratio           & $0.05$         & $0.05$ \\
GKD $\lambda$ & $0.5$  & --- \\
GKD $\beta$        & $0.3$ & --- \\
\bottomrule
\end{tabular}
\caption{Training hyperparameters used.}
\label{tab:hp_recovery}
\end{table}

\begin{table}[t]
\centering
\begin{tabular}{ll}
\toprule
Hyperparameter & Value \\
\midrule
Pruner type            & first-order Taylor \\
Granularity            & block-wise \\
Pruned blocks          & all layers \\
Iterative steps        & $1$ (one-shot) \\
Group aggregation      & sum \\
Pruning scope          & local (per-block) \\
Calibration examples   & $32$ \\
Calibration seq. length & $128$ \\
Seed                   & $42$ \\
\midrule
\multicolumn{2}{l}{\emph{GOLD loss}} \\
JSD $\beta$            & $0.5$ \\
Softmax temperature    & $0.9$ \\
Cross-tokenizer (ULD)  & \textsf{True} \\
\bottomrule
\end{tabular}
\caption{Pruning hyperparameters used. The default calibration loss is CLM. The GOLD loss (bottom block) use the Gemma 3 Instruct 12B as the teacher.}
\label{tab:hp_pruning}
\end{table}

\subsection{Response-Failure Analysis}
\label{app:response-failure}

A drop in task accuracy can arise from two very different failure modes. The model may produces a coherent but \emph{incorrect} answer, or it fails to produce a coherent, on-prompt response at all (degenerate, repetitive, or non-responsive text). The headline metrics conflate the two. To separate them, we analyze the per-sample generations that the LM Evaluation Harness already records (\texttt{samples\_*.jsonl}); this requires no additional inference and reuses the same checkpoints evaluated in the main text.

For each example we keep the raw generation and flag it as \emph{degenerate} with a purely lexical detector: the response is degenerate if its \texttt{zlib} compression ratio falls below $0.30$ (highly repetitive, low-entropy text) \emph{or} its most frequent trigram covers more than $50\%$ of the response (a generation stuck in a loop). Both thresholds are calibrated against the dense model, whose degenerate rate is $0.3\%$ on GSM8K. We then decompose every GSM8K item into three mutually exclusive outcomes: \emph{correct}, \emph{wrong but coherent}, and \emph{degenerate}. The same detector is applied to the free-form IFEval responses. We restrict the figures to the prune-only models,
which isolate the damage caused by pruning before any recovery.

\begin{figure*}[t]
    \centering
    \includegraphics[width=\textwidth]{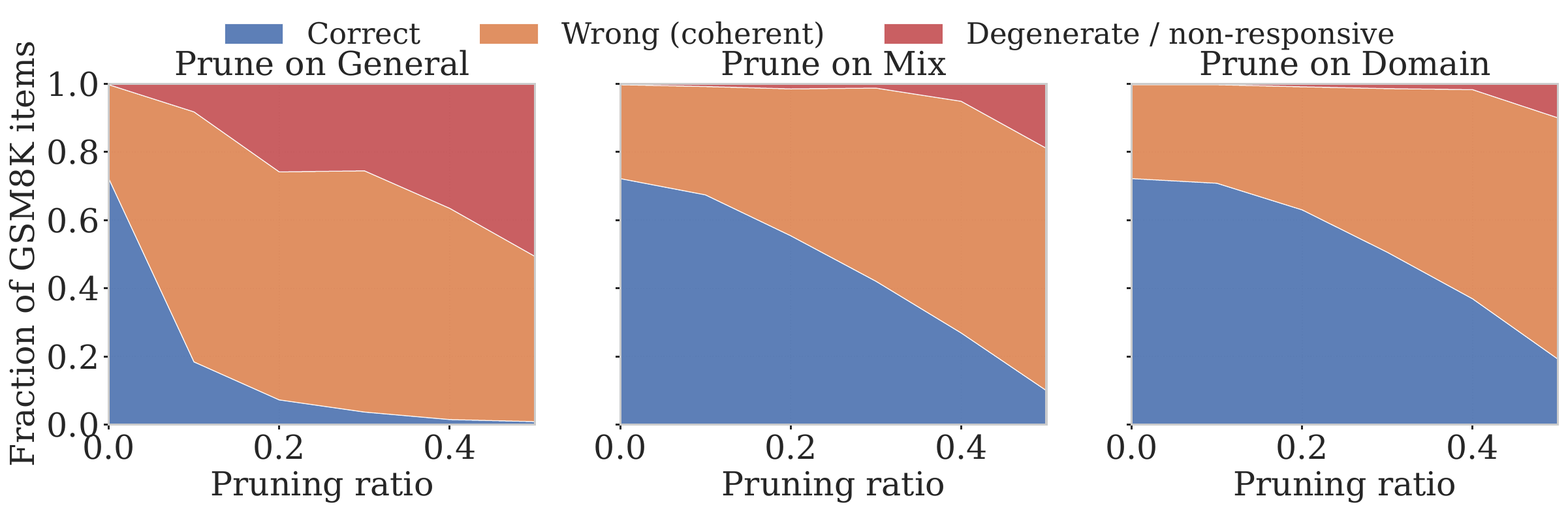}
    \caption{GSM8K outcome decomposition for prune-only models as a function of
    pruning ratio, faceted by the data used for pruning. Each item is classified
    as correct, wrong but coherent, or degenerate / non-responsive. The degenerate
    share grows with sparsity and appears under every pruning-data kind, including
    in-domain pruning.}
    \label{fig:failure_decomp}
\end{figure*}

\begin{figure*}[t]
    \centering
    \includegraphics[width=\textwidth]{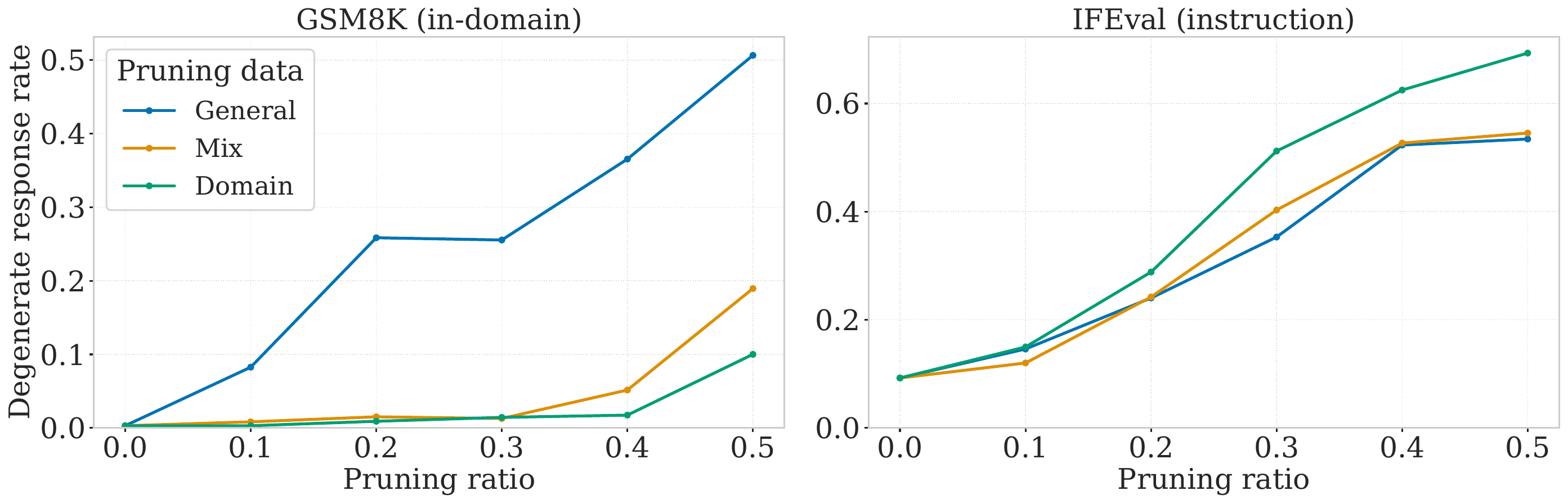}
    \caption{Degenerate-response rate for prune-only models versus pruning ratio,
    on the in-domain task (GSM8K, left) and the instruction-following task
    (IFEval, right), with one line per pruning-data kind. On IFEval the curves are
    nearly independent of the pruning data, indicating a domain-agnostic collapse
    in the model's ability to follow the prompt.}
    \label{fig:degeneration}
\end{figure*}

On in-domain task, a large and growing share of the ``loss'' is not wrong reasoning but non-responses: degenerate outputs rise from $0.3\%$ on the dense model to $27\%$ of all GSM8K items at $50\%$ sparsity, and from $7\%$ to $30\%$ measured as a share of failures (Figure~\ref{fig:failure_decomp}). Crucially, the instruction-following collapse is essentially \emph{independent of the pruning data}. For IFEval the three pruning-data curves overlap and all rise to $0.5$--$0.7$ by $50\%$ sparsity, whereas the GSM8K degeneration rate is data-dependent (general-pruning $0.51$ versus domain-pruning $0.10$ at $50\%$; Figure~\ref{fig:degeneration}). The two collapses co-occur across configurations (Spearman $\rho=0.44$ between the GSM8K and IFEval degeneration rates).

Example degenerate samples:

\begin{quote}\ttfamily
\textbf{Dense:} Quinn is 30 years old. Jane is 3 years younger than Quinn, so Jane
is 30 - 3 = 27 years old. Trent is 5 years older than Jane, so Trent is 27 + 5 =
32 years old. \#\#\#\# 32

\smallskip
\textbf{Pruned (50\%):} Trent is 30 + 3 + 3 + 3 + 5 + 3 + 3 + 3 + 3 = 9 years old.
Trent is 30 + 3 + 3 + 3 + 5 + 3 + 3 + 3 + 3 + 3 + 9 = 9 years old. \#\#\#\# 9
\end{quote}

These observations indicate that the measured in-domain degradation overstates true reasoning loss. A substantial fraction reflects the degradation of the model's ability to respond coherently and follow the prompt, rather than lost task knowledge. This complements the fragility result (\S\ref{sec:fragility}) and supports targeting recovery at instruction-style data
(\S\ref{sec:recovery}), since the failure that recovery must repair is largely one of form, not knowledge. We note that the detector is a lexical heuristic and does not judge semantic quality, and the IFEval degenerate rate has a non-zero floor on the dense model ($9.2\%$), because legitimately structured
answers (lists, JSON) compress well. GSM8K decomposition in the other hand is cleaner since dense baseline is $0.3\%$.

\begin{figure*}[t]
    \centering
    \includegraphics[width=\textwidth]{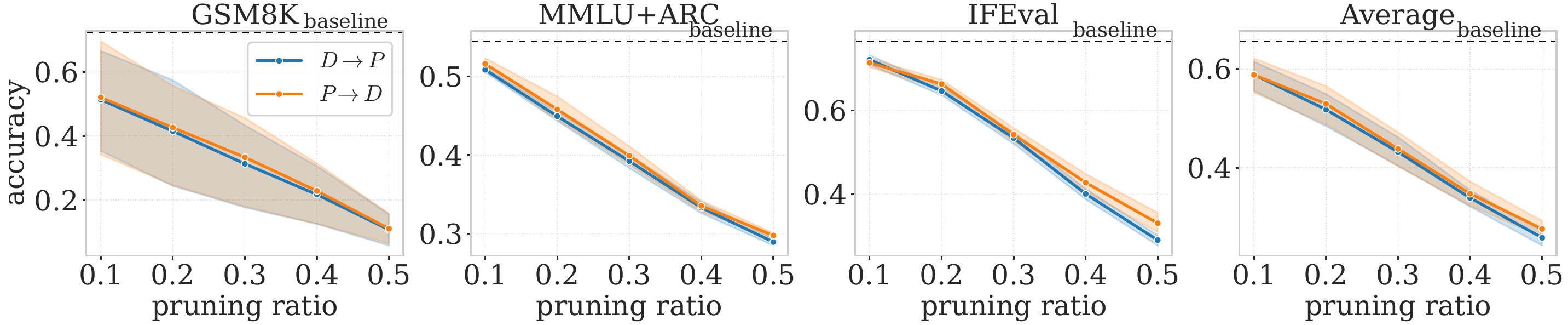}
    \caption{Evaluation performance isolating two major sequences showing little to no effect on final performance.}
    \label{fig:sequences_domain_general}
\end{figure*}

\begin{figure*}[t]
    \centering
    \includegraphics[width=\textwidth]{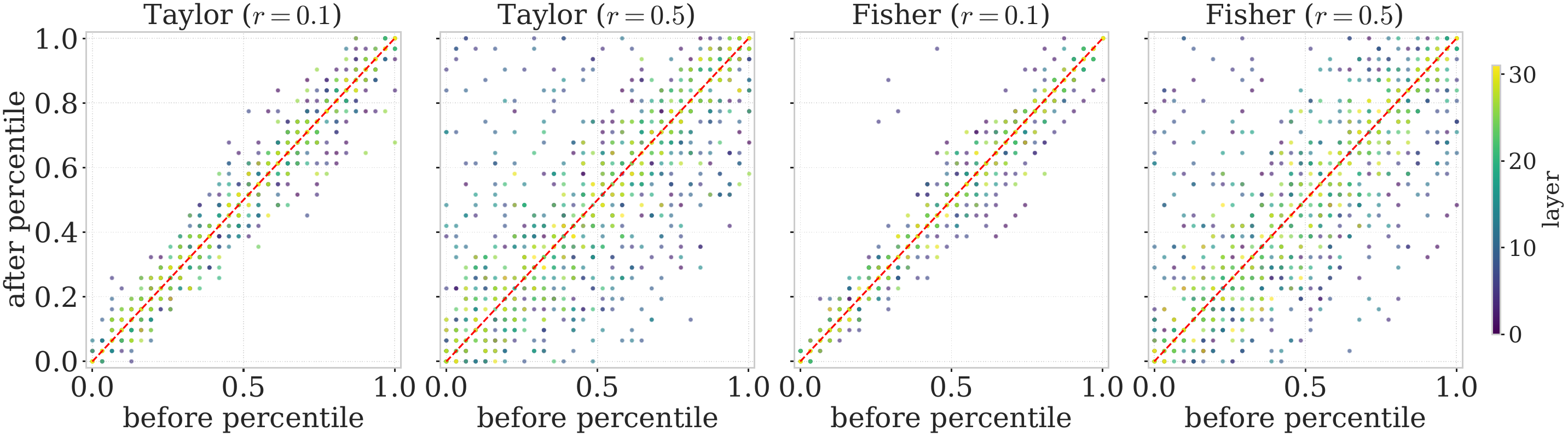}
    \caption{Before- versus after-pruning importance percentile of each surviving
    attention head, colored by layer, for the Taylor and Fisher importance criteria
    at pruning ratios $r=0.1$ and $r=0.5$. The diagonal marks unchanged rank.}
    \label{fig:scatter_head_1x4}
\end{figure*}

\begin{figure*}[t]
    \centering
    \includegraphics[width=\textwidth]{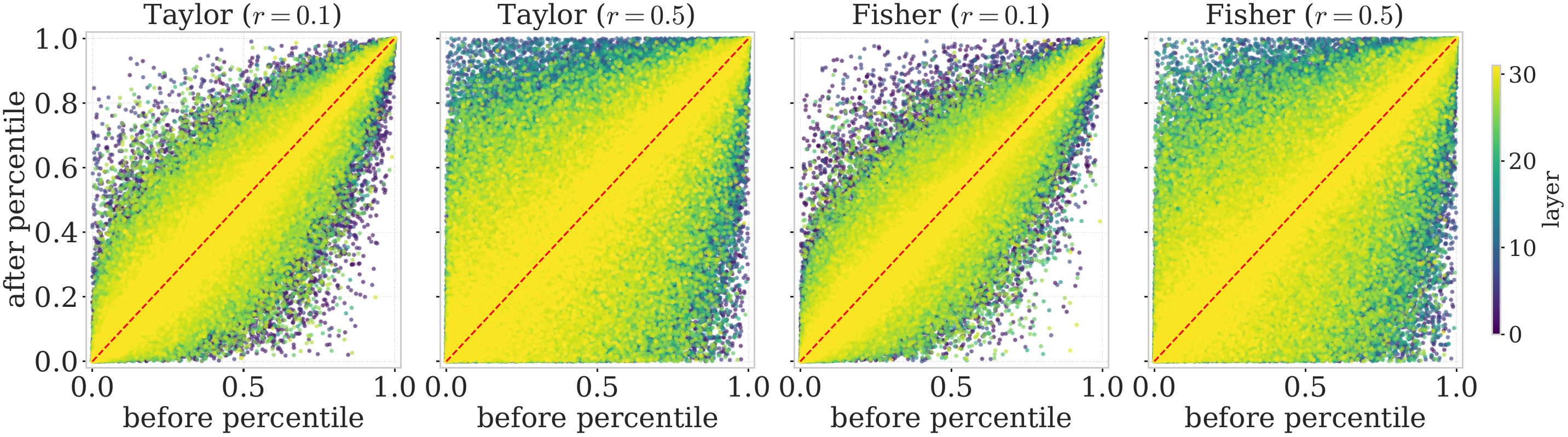}
    \caption{Before- versus after-pruning importance percentile of each surviving
    MLP channel, colored by layer, for the Taylor and Fisher importance criteria
    at pruning ratios $r=0.1$ and $r=0.5$. The diagonal marks unchanged rank.}
    \label{fig:scatter_channel_1x4}
\end{figure*}

\begin{figure*}[t]
    \centering
    \includegraphics[width=\textwidth]{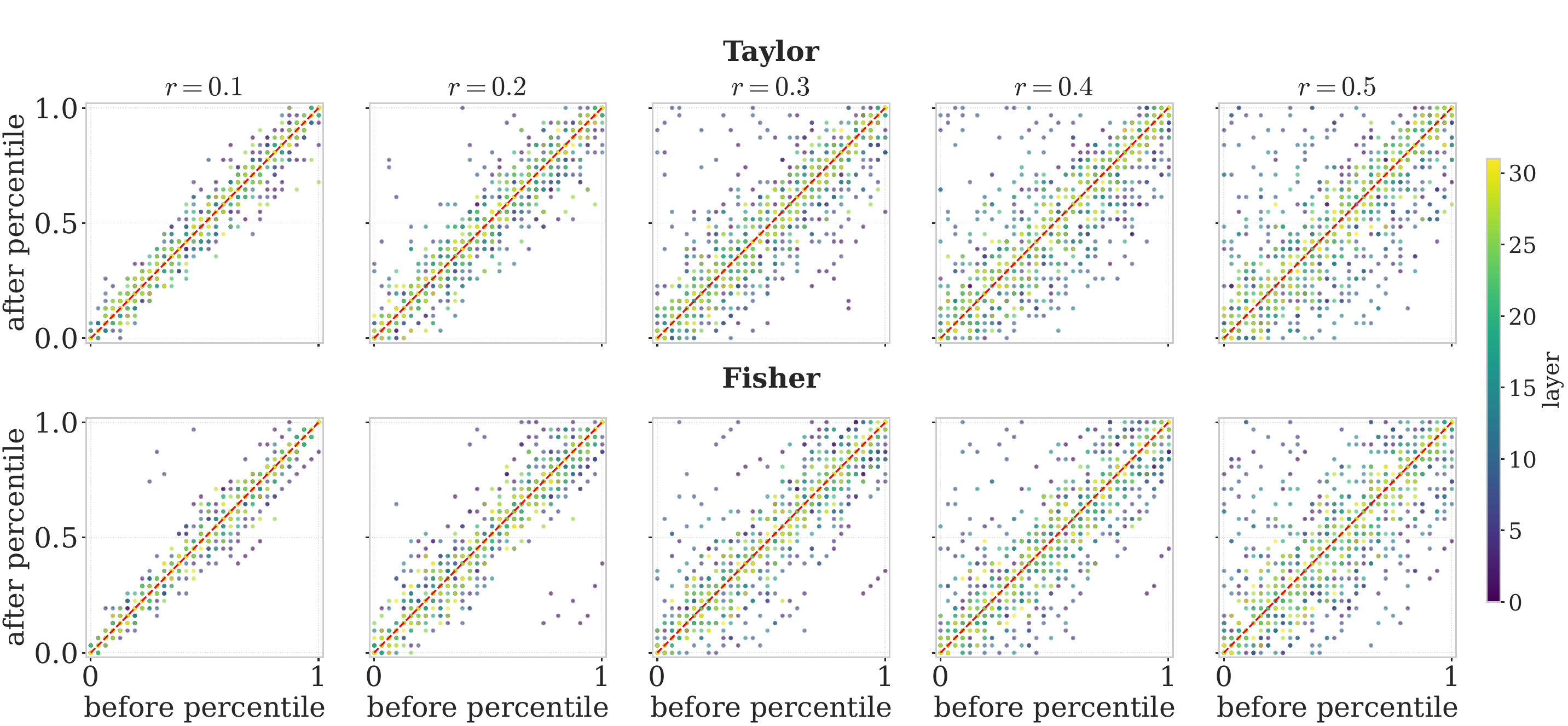}
    \caption{Before- versus after-pruning importance percentile of surviving
    attention heads, colored by layer, for the Taylor and Fisher criteria across
    pruning ratios $r=0.1$ to $r=0.5$.}
    \label{fig:scatter_head_2x5}
\end{figure*}

\begin{figure*}[t]
    \centering
    \includegraphics[width=\textwidth]{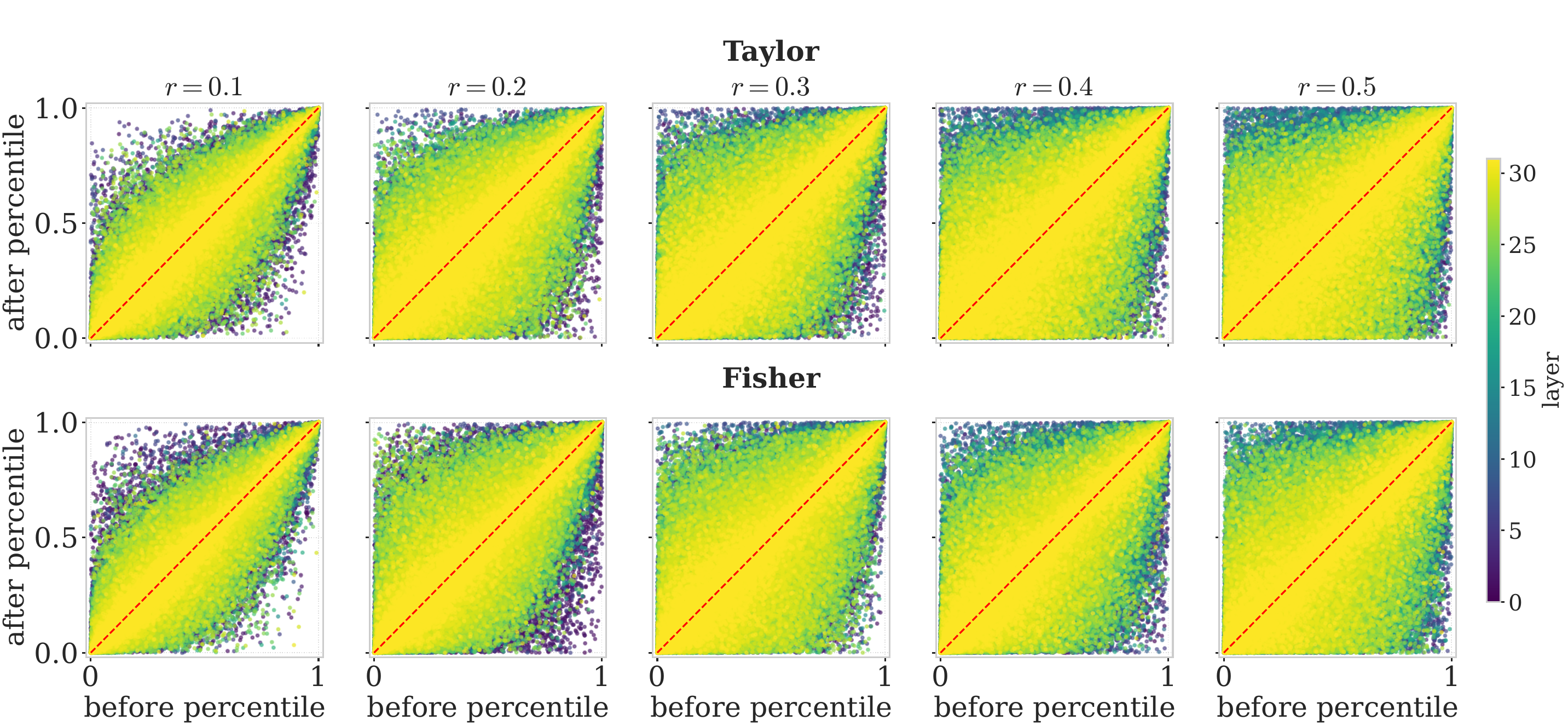}
    \caption{Before- versus after-pruning importance percentile of surviving
    MLP channels, colored by layer, for the Taylor and Fisher criteria across
    pruning ratios $r=0.1$ to $r=0.5$.}
    \label{fig:scatter_channel_2x5}
\end{figure*}

\begin{figure*}[t]
    \centering
    \includegraphics[width=\textwidth]{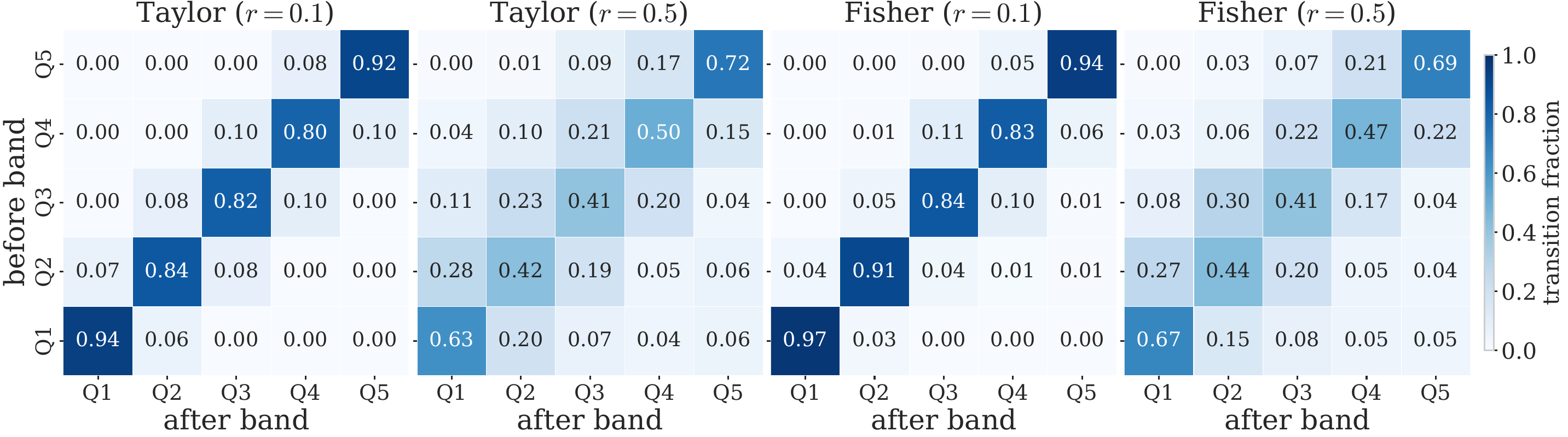}
    \caption{Band-transition matrices of attention-head importance before versus
    after pruning, shown for the Taylor and Fisher importance criteria at pruning
    ratios $r=0.1$ and $r=0.5$. Each cell gives the fraction of heads moving from
    a before-pruning importance quintile (row) to an after-pruning quintile (column).}
    \label{fig:transition_head_1x4}
\end{figure*}


\begin{figure*}[t]
    \centering
    \includegraphics[width=\textwidth]{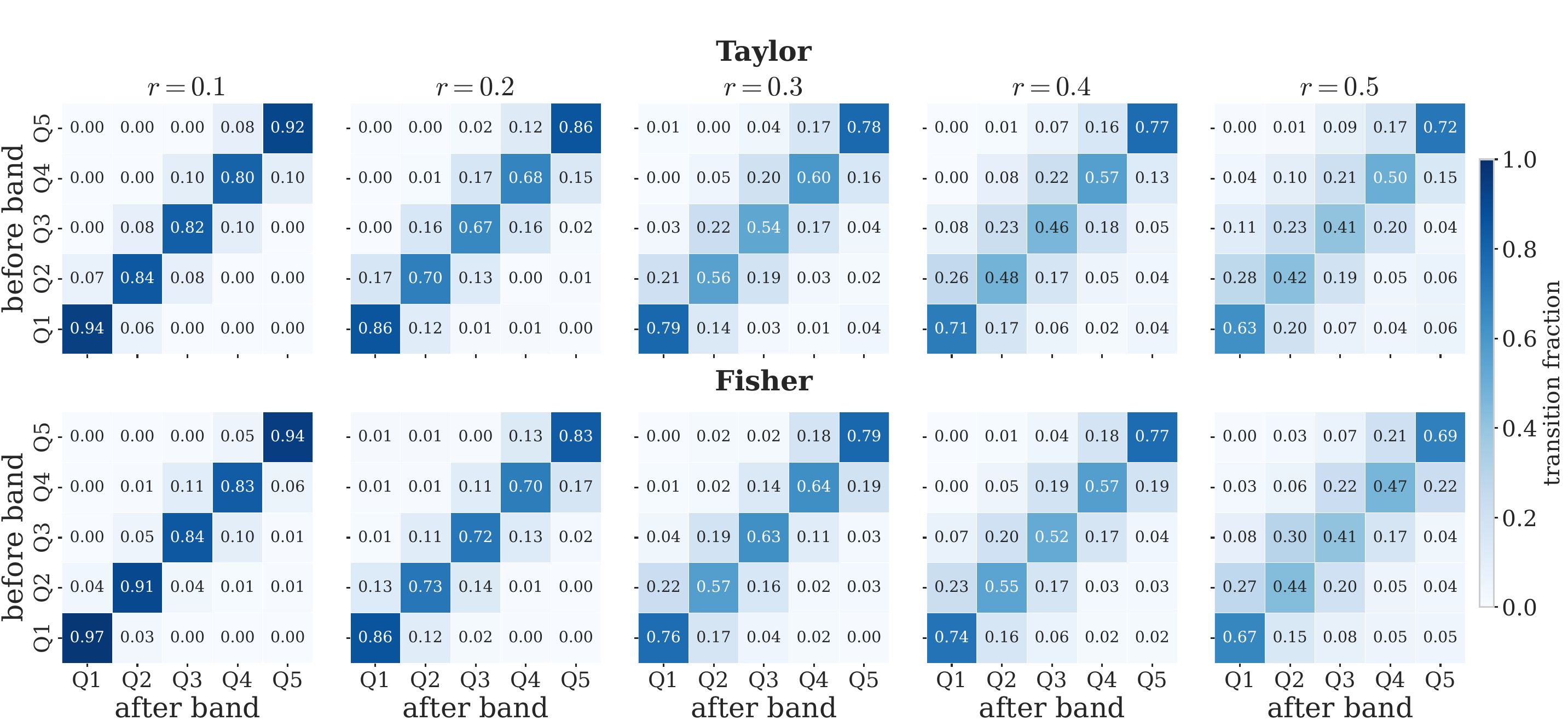}
    \caption{Attention-head importance band-transition matrices for the Taylor
    (top row) and Fisher (bottom row) criteria across pruning ratios
    $r=0.1$ to $r=0.5$ (columns).}
    \label{fig:transition_head_2x5}
\end{figure*}

\begin{figure*}[t]
    \centering
    \includegraphics[width=\textwidth]{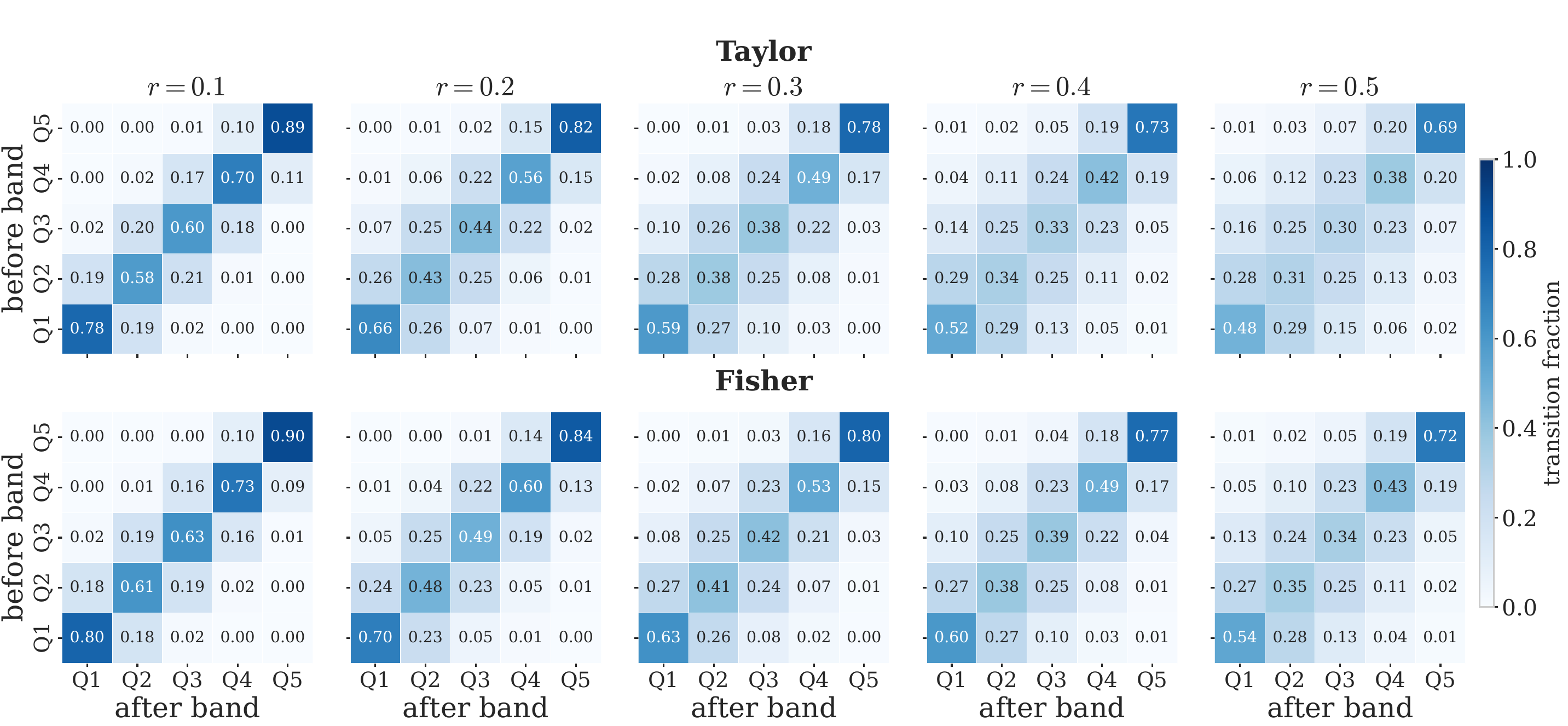}
    \caption{MLP-channel importance band-transition matrices for the Taylor
    (top row) and Fisher (bottom row) criteria across pruning ratios
    $r=0.1$ to $r=0.5$ (columns).}
    \label{fig:transition_channel_2x5}
\end{figure*}

\begin{figure*}[t]
    \centering
    \includegraphics[width=\textwidth]{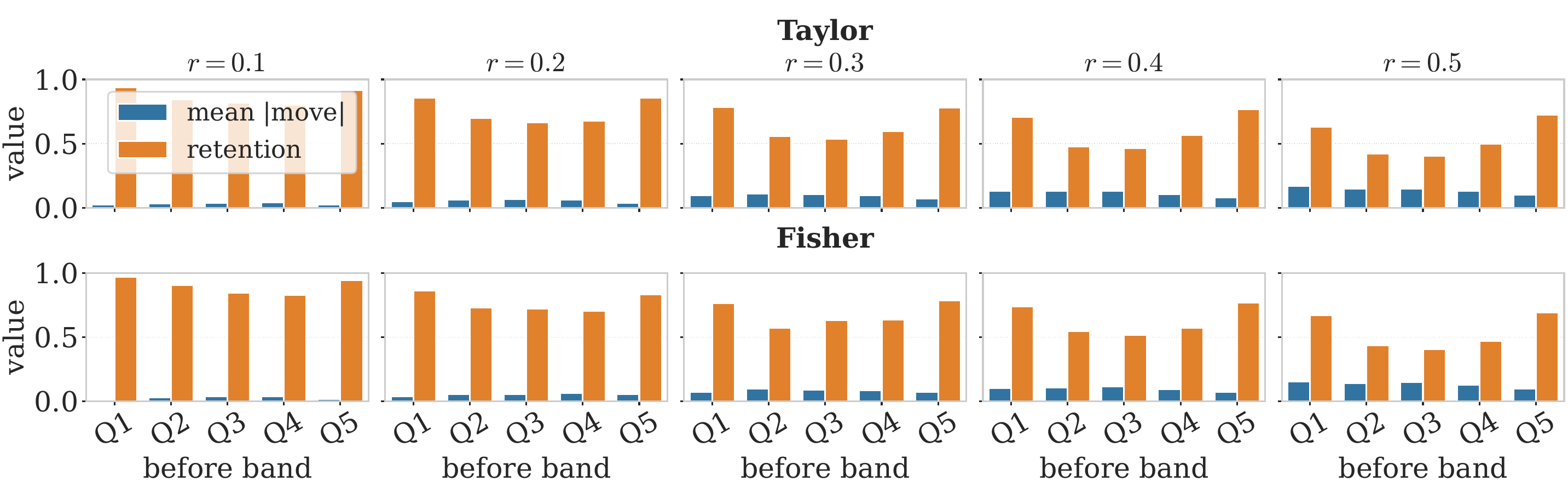}
    \caption{Per-quintile mean absolute importance movement and retention for
    attention heads, for the Taylor and Fisher criteria across pruning ratios
    $r=0.1$ to $r=0.5$.}
    \label{fig:bands_head_2x5}
\end{figure*}

\begin{figure*}[t]
    \centering
    \includegraphics[width=\textwidth]{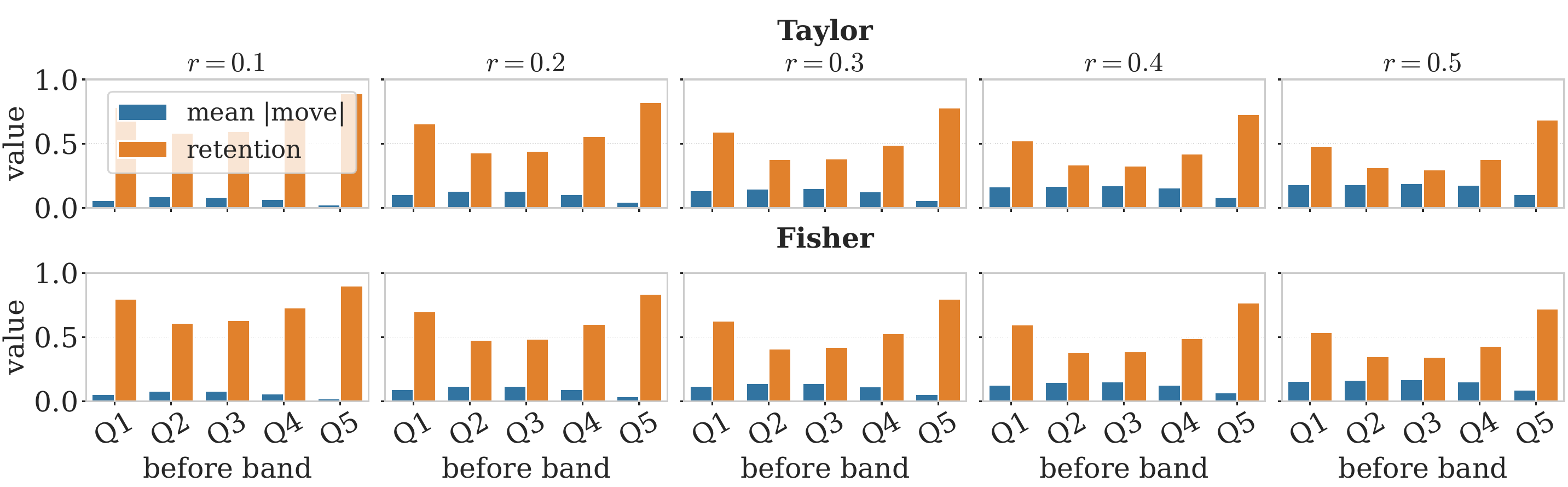}
    \caption{Per-quintile mean absolute importance movement and retention for
    MLP channels, for the Taylor and Fisher criteria across pruning ratios
    $r=0.1$ to $r=0.5$.}
    \label{fig:bands_channel_2x5}
\end{figure*}




\begin{figure*}[t]
    \centering
    \includegraphics[width=\textwidth]{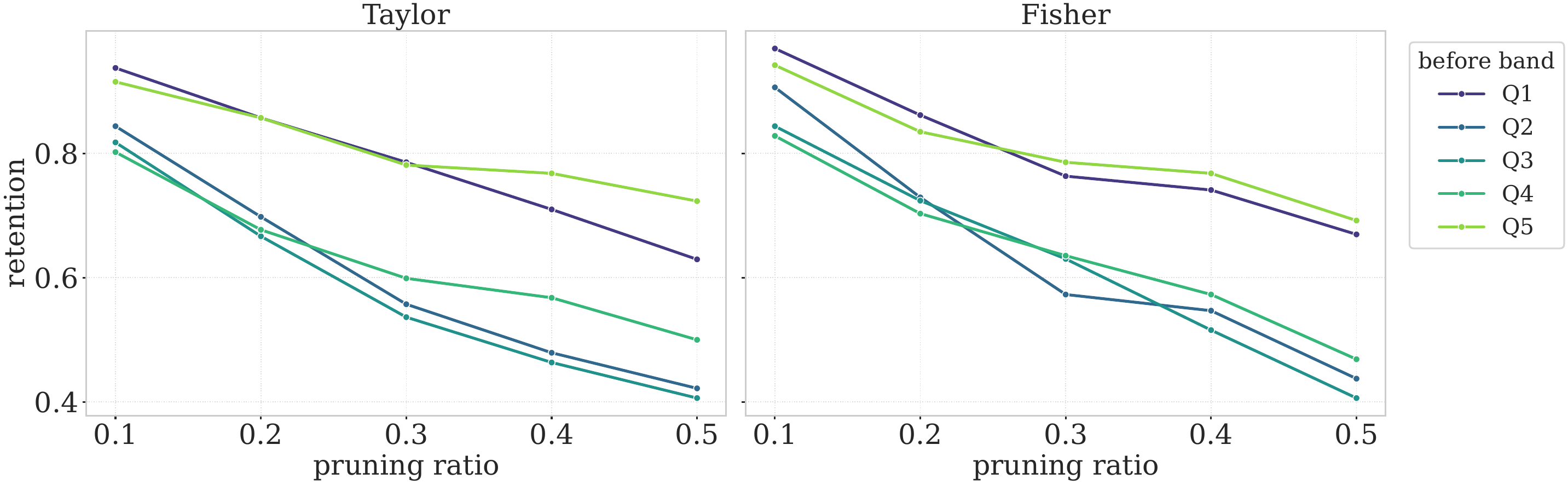}
    \caption{Per-quintile retention of attention-head importance as a function
    of pruning ratio, faceted by importance criterion (Taylor and Fisher), with
    one line per before-pruning quintile.}
    \label{fig:retention_head}
\end{figure*}

\begin{figure*}[t]
    \centering
    \includegraphics[width=\textwidth]{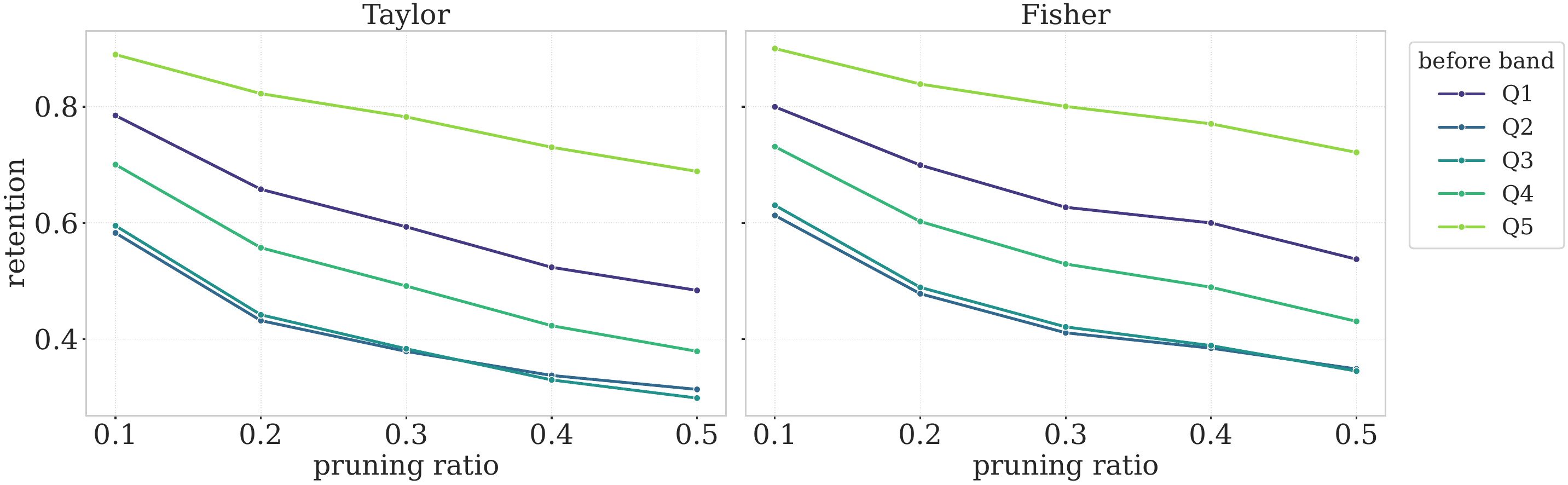}
    \caption{Per-quintile retention of MLP-channel importance as a function of
    pruning ratio, faceted by importance criterion (Taylor and Fisher), with one
    line per before-pruning quintile.}
    \label{fig:retention_channel}
\end{figure*}

\begin{figure*}[t]
    \centering
    \includegraphics[width=\textwidth]{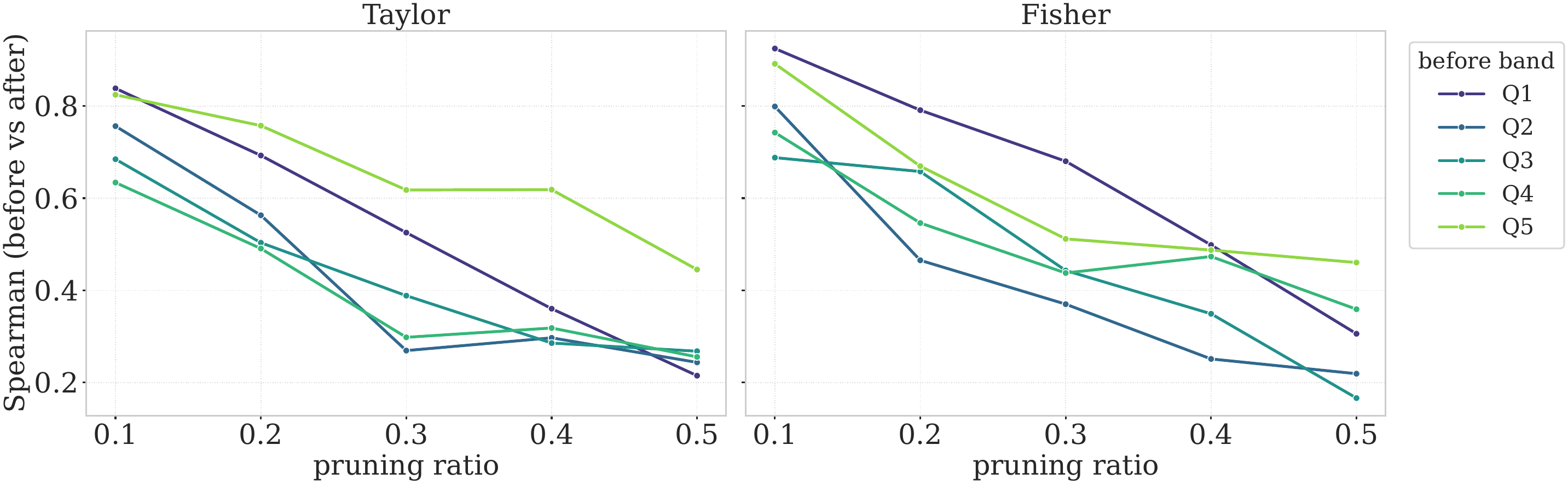}
    \caption{Spearman rank correlation between before- and after-pruning
    attention-head importance as a function of pruning ratio, resolved per
    before-pruning quintile and faceted by importance criterion.}
    \label{fig:spearman_head}
\end{figure*}

\begin{figure*}[t]
    \centering
    \includegraphics[width=\textwidth]{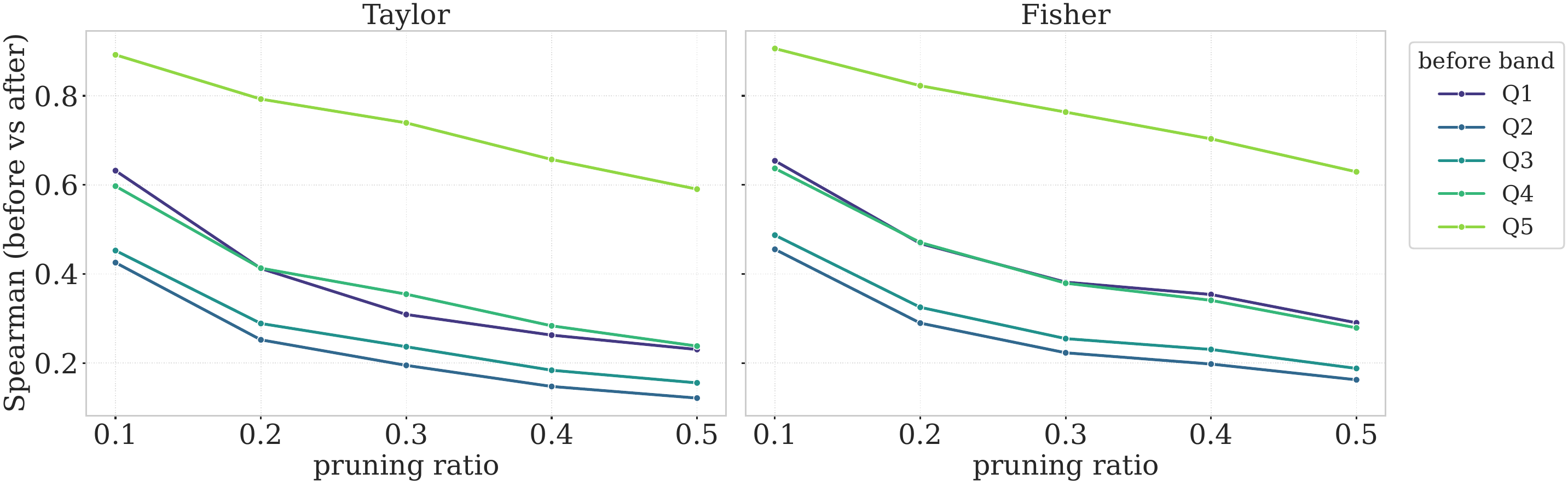}
    \caption{Spearman rank correlation between before- and after-pruning
    MLP-channel importance as a function of pruning ratio, resolved per
    before-pruning quintile and faceted by importance criterion.}
    \label{fig:spearman_channel}
\end{figure*}

\subsection{Full Results}
\label{app:full-results}

For completeness, Table~\ref{tab:full_results} reports the aggregate metrics for every configuration in the main sweep, and Table~\ref{tab:full_alignment} reports the loss-alignment runs. We report only whole metrics without its subcategories.

\begin{table*}[t]
\centering
\small
\begin{tabular}{lll rrr @{\hskip 2em} lll rrr}
\toprule
Loss & Recov & $r$ & GSM8K & Gen & Avg & Loss & Recov & $r$ & GSM8K & Gen & Avg \\
\midrule
CLM & KD & 0.1 & 0.171 & 0.519 & 0.525 & GOLD & KD & 0.1 & 0.208 & 0.476 & 0.483 \\
CLM & KD & 0.2 & 0.074 & 0.442 & 0.451 & GOLD & KD & 0.2 & 0.108 & 0.443 & 0.454 \\
CLM & KD & 0.3 & 0.032 & 0.387 & 0.395 & GOLD & KD & 0.3 & 0.049 & 0.364 & 0.375 \\
CLM & KD & 0.4 & 0.024 & 0.322 & 0.326 & GOLD & KD & 0.4 & 0.027 & 0.310 & 0.320 \\
CLM & KD & 0.5 & 0.017 & 0.296 & 0.302 & GOLD & KD & 0.5 & 0.026 & 0.288 & 0.292 \\
CLM & SFT & 0.1 & 0.185 & 0.506 & 0.515 & GOLD & SFT & 0.1 & 0.227 & 0.497 & 0.507 \\
CLM & SFT & 0.2 & 0.076 & 0.431 & 0.443 & GOLD & SFT & 0.2 & 0.104 & 0.435 & 0.451 \\
CLM & SFT & 0.3 & 0.041 & 0.367 & 0.373 & GOLD & SFT & 0.3 & 0.048 & 0.365 & 0.381 \\
CLM & SFT & 0.4 & 0.018 & 0.321 & 0.329 & GOLD & SFT & 0.4 & 0.035 & 0.311 & 0.319 \\
CLM & SFT & 0.5 & 0.014 & 0.284 & 0.286 & GOLD & SFT & 0.5 & 0.021 & 0.275 & 0.279 \\
CLM & none & 0.1 & 0.187 & 0.505 & 0.514 & GOLD & none & 0.1 & 0.220 & 0.498 & 0.508 \\
CLM & none & 0.2 & 0.077 & 0.431 & 0.444 & GOLD & none & 0.2 & 0.106 & 0.435 & 0.451 \\
CLM & none & 0.3 & 0.042 & 0.368 & 0.373 & GOLD & none & 0.3 & 0.049 & 0.366 & 0.381 \\
CLM & none & 0.4 & 0.020 & 0.322 & 0.330 & GOLD & none & 0.4 & 0.037 & 0.312 & 0.319 \\
CLM & none & 0.5 & 0.015 & 0.283 & 0.285 & GOLD & none & 0.5 & 0.022 & 0.276 & 0.279 \\
\bottomrule
\end{tabular}
\caption{Loss-alignment runs. Loss is the pruning calibration objective (CLM or GOLD), Recov is the recovery method (none = pruning-only, SFT, or KD/GOLD distillation), $r$ is the pruning ratio. IFEval was not evaluated for these runs. Rows continue in the right-hand block.}
\label{tab:full_alignment}
\end{table*}

\begin{table*}[t]
\small
\centering
\setlength{\tabcolsep}{4pt}
\begin{tabular}{llll rrrr @{\hskip 1em} llll rrrr}
\toprule
Seq & $P$ & $D$ & $r$ & GSM8K & Gen & IFE & Avg & Seq & $P$ & $D$ & $r$ & GSM8K & Gen & IFE & Avg \\
\midrule
base & -- & -- & -- & 0.723 & 0.545 & 0.764 & 0.656 & $D{\to}P$ & Mix & Gen & 0.2 & 0.552 & 0.461 & 0.663 & 0.551 \\
$P$ & Dom & -- & 0.1 & 0.708 & 0.511 & 0.709 & 0.622 & $D{\to}P$ & Mix & Gen & 0.3 & 0.420 & 0.395 & 0.521 & 0.448 \\
$P$ & Dom & -- & 0.2 & 0.630 & 0.453 & 0.644 & 0.561 & $D{\to}P$ & Mix & Gen & 0.4 & 0.281 & 0.329 & 0.403 & 0.349 \\
$P$ & Dom & -- & 0.3 & 0.506 & 0.399 & 0.546 & 0.476 & $D{\to}P$ & Mix & Gen & 0.5 & 0.112 & 0.292 & 0.288 & 0.259 \\
$P$ & Dom & -- & 0.4 & 0.369 & 0.343 & 0.377 & 0.361 & $D{\to}P$ & Mix & Mix & 0.1 & 0.663 & 0.509 & 0.741 & 0.625 \\
$P$ & Dom & -- & 0.5 & 0.196 & 0.295 & 0.309 & 0.283 & $D{\to}P$ & Mix & Mix & 0.2 & 0.542 & 0.457 & 0.645 & 0.540 \\
$P$ & Gen & -- & 0.1 & 0.187 & 0.505 & 0.702 & 0.521 & $D{\to}P$ & Mix & Mix & 0.3 & 0.412 & 0.397 & 0.521 & 0.449 \\
$P$ & Gen & -- & 0.2 & 0.077 & 0.431 & 0.642 & 0.448 & $D{\to}P$ & Mix & Mix & 0.4 & 0.258 & 0.332 & 0.399 & 0.345 \\
$P$ & Gen & -- & 0.3 & 0.042 & 0.368 & 0.532 & 0.370 & $D{\to}P$ & Mix & Mix & 0.5 & 0.108 & 0.292 & 0.296 & 0.262 \\
$P$ & Gen & -- & 0.4 & 0.020 & 0.322 & 0.403 & 0.300 & $P{\to}D$ & Dom & Dom & 0.1 & 0.688 & 0.504 & 0.717 & 0.615 \\
$P$ & Gen & -- & 0.5 & 0.015 & 0.283 & 0.281 & 0.234 & $P{\to}D$ & Dom & Dom & 0.2 & 0.611 & 0.451 & 0.671 & 0.565 \\
$P$ & Mix & -- & 0.1 & 0.675 & 0.508 & 0.735 & 0.623 & $P{\to}D$ & Dom & Dom & 0.3 & 0.518 & 0.396 & 0.555 & 0.476 \\
$P$ & Mix & -- & 0.2 & 0.556 & 0.459 & 0.667 & 0.554 & $P{\to}D$ & Dom & Dom & 0.4 & 0.365 & 0.336 & 0.407 & 0.367 \\
$P$ & Mix & -- & 0.3 & 0.422 & 0.392 & 0.532 & 0.452 & $P{\to}D$ & Dom & Dom & 0.5 & 0.209 & 0.298 & 0.321 & 0.289 \\
$P$ & Mix & -- & 0.4 & 0.270 & 0.326 & 0.408 & 0.349 & $P{\to}D$ & Dom & Gen & 0.1 & 0.707 & 0.504 & 0.711 & 0.613 \\
$P$ & Mix & -- & 0.5 & 0.105 & 0.292 & 0.305 & 0.265 & $P{\to}D$ & Dom & Gen & 0.2 & 0.625 & 0.439 & 0.673 & 0.562 \\
$D$ & -- & Dom & -- & 0.712 & 0.540 & 0.756 & 0.640 & $P{\to}D$ & Dom & Gen & 0.3 & 0.509 & 0.398 & 0.580 & 0.483 \\
$D$ & -- & Gen & -- & 0.720 & 0.550 & 0.748 & 0.651 & $P{\to}D$ & Dom & Gen & 0.4 & 0.370 & 0.341 & 0.474 & 0.393 \\
$D$ & -- & Mix & -- & 0.714 & 0.547 & 0.771 & 0.654 & $P{\to}D$ & Dom & Gen & 0.5 & 0.177 & 0.288 & 0.403 & 0.312 \\
$D{\to}P$ & Dom & Dom & 0.1 & 0.681 & 0.501 & 0.709 & 0.606 & $P{\to}D$ & Dom & Mix & 0.1 & 0.694 & 0.500 & 0.712 & 0.608 \\
$D{\to}P$ & Dom & Dom & 0.2 & 0.612 & 0.445 & 0.666 & 0.562 & $P{\to}D$ & Dom & Mix & 0.2 & 0.630 & 0.438 & 0.668 & 0.560 \\
$D{\to}P$ & Dom & Dom & 0.3 & 0.486 & 0.412 & 0.571 & 0.487 & $P{\to}D$ & Dom & Mix & 0.3 & 0.528 & 0.395 & 0.572 & 0.481 \\
$D{\to}P$ & Dom & Dom & 0.4 & 0.356 & 0.349 & 0.370 & 0.360 & $P{\to}D$ & Dom & Mix & 0.4 & 0.378 & 0.345 & 0.458 & 0.391 \\
$D{\to}P$ & Dom & Dom & 0.5 & 0.199 & 0.291 & 0.304 & 0.282 & $P{\to}D$ & Dom & Mix & 0.5 & 0.198 & 0.298 & 0.376 & 0.309 \\
$D{\to}P$ & Dom & Gen & 0.1 & 0.690 & 0.518 & 0.729 & 0.627 & $P{\to}D$ & Gen & Dom & 0.1 & 0.156 & 0.512 & 0.700 & 0.515 \\
$D{\to}P$ & Dom & Gen & 0.2 & 0.634 & 0.456 & 0.630 & 0.557 & $P{\to}D$ & Gen & Dom & 0.2 & 0.060 & 0.436 & 0.642 & 0.445 \\
$D{\to}P$ & Dom & Gen & 0.3 & 0.506 & 0.402 & 0.553 & 0.478 & $P{\to}D$ & Gen & Dom & 0.3 & 0.024 & 0.373 & 0.531 & 0.370 \\
$D{\to}P$ & Dom & Gen & 0.4 & 0.371 & 0.348 & 0.384 & 0.364 & $P{\to}D$ & Gen & Dom & 0.4 & 0.031 & 0.316 & 0.387 & 0.292 \\
$D{\to}P$ & Dom & Gen & 0.5 & 0.201 & 0.298 & 0.321 & 0.289 & $P{\to}D$ & Gen & Dom & 0.5 & 0.020 & 0.293 & 0.295 & 0.245 \\
$D{\to}P$ & Dom & Mix & 0.1 & 0.691 & 0.513 & 0.729 & 0.624 & $P{\to}D$ & Gen & Gen & 0.1 & 0.168 & 0.521 & 0.682 & 0.514 \\
$D{\to}P$ & Dom & Mix & 0.2 & 0.623 & 0.448 & 0.661 & 0.564 & $P{\to}D$ & Gen & Gen & 0.2 & 0.074 & 0.444 & 0.643 & 0.452 \\
$D{\to}P$ & Dom & Mix & 0.3 & 0.504 & 0.408 & 0.550 & 0.480 & $P{\to}D$ & Gen & Gen & 0.3 & 0.033 & 0.387 & 0.504 & 0.368 \\
$D{\to}P$ & Dom & Mix & 0.4 & 0.372 & 0.346 & 0.385 & 0.366 & $P{\to}D$ & Gen & Gen & 0.4 & 0.019 & 0.322 & 0.396 & 0.295 \\
$D{\to}P$ & Dom & Mix & 0.5 & 0.190 & 0.297 & 0.287 & 0.275 & $P{\to}D$ & Gen & Gen & 0.5 & 0.022 & 0.297 & 0.281 & 0.242 \\
$D{\to}P$ & Gen & Dom & 0.1 & 0.182 & 0.503 & 0.715 & 0.524 & $P{\to}D$ & Gen & Mix & 0.1 & 0.168 & 0.517 & 0.702 & 0.520 \\
$D{\to}P$ & Gen & Dom & 0.2 & 0.083 & 0.436 & 0.632 & 0.446 & $P{\to}D$ & Gen & Mix & 0.2 & 0.071 & 0.440 & 0.631 & 0.445 \\
$D{\to}P$ & Gen & Dom & 0.3 & 0.036 & 0.373 & 0.549 & 0.377 & $P{\to}D$ & Gen & Mix & 0.3 & 0.029 & 0.381 & 0.499 & 0.363 \\
$D{\to}P$ & Gen & Dom & 0.4 & 0.016 & 0.322 & 0.428 & 0.309 & $P{\to}D$ & Gen & Mix & 0.4 & 0.024 & 0.323 & 0.394 & 0.295 \\
$D{\to}P$ & Gen & Dom & 0.5 & 0.019 & 0.278 & 0.294 & 0.239 & $P{\to}D$ & Gen & Mix & 0.5 & 0.027 & 0.294 & 0.311 & 0.252 \\
$D{\to}P$ & Gen & Gen & 0.1 & 0.190 & 0.509 & 0.703 & 0.524 & $P{\to}D$ & Mix & Dom & 0.1 & 0.695 & 0.535 & 0.726 & 0.635 \\
$D{\to}P$ & Gen & Gen & 0.2 & 0.083 & 0.440 & 0.621 & 0.444 & $P{\to}D$ & Mix & Dom & 0.2 & 0.581 & 0.498 & 0.672 & 0.576 \\
$D{\to}P$ & Gen & Gen & 0.3 & 0.031 & 0.377 & 0.504 & 0.362 & $P{\to}D$ & Mix & Dom & 0.3 & 0.444 & 0.419 & 0.549 & 0.467 \\
$D{\to}P$ & Gen & Gen & 0.4 & 0.018 & 0.323 & 0.397 & 0.298 & $P{\to}D$ & Mix & Dom & 0.4 & 0.299 & 0.345 & 0.413 & 0.355 \\
$D{\to}P$ & Gen & Gen & 0.5 & 0.020 & 0.281 & 0.270 & 0.231 & $P{\to}D$ & Mix & Dom & 0.5 & 0.119 & 0.301 & 0.326 & 0.276 \\
$D{\to}P$ & Gen & Mix & 0.1 & 0.202 & 0.506 & 0.686 & 0.519 & $P{\to}D$ & Mix & Gen & 0.1 & 0.706 & 0.526 & 0.737 & 0.636 \\
$D{\to}P$ & Gen & Mix & 0.2 & 0.080 & 0.441 & 0.640 & 0.451 & $P{\to}D$ & Mix & Gen & 0.2 & 0.600 & 0.487 & 0.680 & 0.579 \\
$D{\to}P$ & Gen & Mix & 0.3 & 0.033 & 0.374 & 0.518 & 0.366 & $P{\to}D$ & Mix & Gen & 0.3 & 0.466 & 0.423 & 0.543 & 0.470 \\
$D{\to}P$ & Gen & Mix & 0.4 & 0.020 & 0.320 & 0.413 & 0.304 & $P{\to}D$ & Mix & Gen & 0.4 & 0.284 & 0.343 & 0.457 & 0.367 \\
$D{\to}P$ & Gen & Mix & 0.5 & 0.019 & 0.281 & 0.254 & 0.224 & $P{\to}D$ & Mix & Gen & 0.5 & 0.107 & 0.304 & 0.343 & 0.280 \\
$D{\to}P$ & Mix & Dom & 0.1 & 0.647 & 0.509 & 0.730 & 0.618 & $P{\to}D$ & Mix & Mix & 0.1 & 0.705 & 0.528 & 0.732 & 0.635 \\
$D{\to}P$ & Mix & Dom & 0.2 & 0.528 & 0.461 & 0.653 & 0.541 & $P{\to}D$ & Mix & Mix & 0.2 & 0.585 & 0.491 & 0.683 & 0.579 \\
$D{\to}P$ & Mix & Dom & 0.3 & 0.393 & 0.394 & 0.516 & 0.440 & $P{\to}D$ & Mix & Mix & 0.3 & 0.453 & 0.421 & 0.545 & 0.466 \\
$D{\to}P$ & Mix & Dom & 0.4 & 0.268 & 0.330 & 0.428 & 0.355 & $P{\to}D$ & Mix & Mix & 0.4 & 0.290 & 0.347 & 0.464 & 0.374 \\
$D{\to}P$ & Mix & Dom & 0.5 & 0.103 & 0.294 & 0.300 & 0.264 & $P{\to}D$ & Mix & Mix & 0.5 & 0.122 & 0.305 & 0.321 & 0.278 \\
$D{\to}P$ & Mix & Gen & 0.1 & 0.676 & 0.511 & 0.741 & 0.627 &  &  &  &  &  &  &  &  \\
\bottomrule
\end{tabular}
\caption{Full per-configuration results, reporting aggregate metrics only. Seq is the stage order ($D{\to}P$, $P{\to}D$), $P$/$D$ are the pruning/distillation data kinds, $r$ is the pruning ratio. GSM8K is flexible-extract, General is the mean of MMLU and ARC-easy/challenge, IFE is the mean of the four IFEval splits, and Avg is the overall average. Rows continue in the right-hand block.}
\label{tab:full_results}
\end{table*}

\end{document}